%% file: main.tex
\documentclass[11pt]{article}

\usepackage[preprint]{acl}

\usepackage{times}
\usepackage{latexsym}

\usepackage[T1]{fontenc}

\usepackage[utf8]{inputenc}

\usepackage{microtype}

\usepackage{inconsolata}

\usepackage{graphicx}
\usepackage{multirow}
\usepackage[table]{xcolor}

\input{_packages}

\title{Enabling Labelling of Speech Translation Errors}
\title{Automatic Labelling of Speech Translation Errors}

\newcommand{\authorsep}{\quad}

\author{
 \textbf{Dominik Macháček}\textsuperscript{1,2}\authorsep
 \textbf{Maike Züfle}\textsuperscript{3}\authorsep
 \textbf{Ondrej Klejch}\textsuperscript{2}
\\
 \textsuperscript{1}Charles University\authorsep
 \textsuperscript{2}University of Edinburgh\authorsep
 \textsuperscript{3}Karlsruhe Institute of Technology
\\
 \small{\texttt{machacek@ufal.mff.cuni.cz}}
}

\begin{document}
\maketitle
\input{content}
\bibliography{stel-bibs,old-bibs,anthology-1}

\appendix
\input{appendix}

\end{document}

%% file: _packages.tex
\usepackage{microtype}
\usepackage{booktabs,arydshln}
\usepackage{array}
\usepackage{ragged2e}
 \usepackage{amsmath}
 \usepackage{threeparttable}
 \usepackage{multicol}
 \usepackage{multirow}
 \usepackage{ccicons}
 \usepackage{chngcntr}

 \makeatletter
\def\adl@drawiv#1#2#3{%
        \hskip.5\tabcolsep
        \xleaders#3{#2.5\@tempdimb #1{1}#2.5\@tempdimb}%
                #2\z@ plus1fil minus1fil\relax
        \hskip.5\tabcolsep}
\newcommand{\cdashlinelr}[1]{%
  \noalign{\vskip\aboverulesep
           \global\let\@dashdrawstore\adl@draw
           \global\let\adl@draw\adl@drawiv}
  \cdashline{#1}
  \noalign{\global\let\adl@draw\@dashdrawstore
           \vskip\belowrulesep}}
\makeatother

\makeatletter
\def\adl@drawiv#1#2#3{%
  \hskip.5\tabcolsep
  {\color[gray]{0.7} 
  \xleaders#3{#2.5\@tempdimb #1{1}#2.5\@tempdimb}%
      #2\z@ plus1fil minus1fil\relax}%
  \hskip.5\tabcolsep}
\newcommand{\cdashlinelrg}[1]{%
  \noalign{\vskip\aboverulesep
           \global\let\@dashdrawstore\adl@draw
           \global\let\adl@draw\adl@drawiv}
  \cdashline{#1}
  \noalign{\global\let\adl@draw\@dashdrawstore
           \vskip\belowrulesep}}
\makeatother

\usepackage{pifont}
\usepackage{fontawesome5}

\definecolor{richgreen}{RGB}{102, 204, 102}
\definecolor{richred}{RGB}{255, 99, 71}

\definecolor{richorange}{RGB}{255, 165, 0}

\definecolor{ctxorange}{RGB}{217,95,2} 
\definecolor{ctxteal}{RGB}{27,158,119} 
\definecolor{coolgray}{RGB}{160,160,170}

\definecolor{slateblue}{RGB}{100,140,200}

\definecolor{augpurple}{RGB}{117,107,177}

\usepackage[most]{tcolorbox}   
\usepackage{siunitx}           
\usepackage{colortbl} 

\newcolumntype{R}{p{0.05cm}} 

\newcolumntype{P}[1]{>{\RaggedRight\arraybackslash}p{#1}}
\usepackage{makecell}
\newcolumntype{C}[1]{>{\centering\arraybackslash}p{#1}}
\newcolumntype{D}[1]{>{\raggedright\arraybackslash}p{#1}}

\usepackage{amsmath}
\usepackage{amsfonts}
\usepackage{inconsolata}

\usepackage{graphicx}

\usepackage{blindtext}

\usepackage{afterpage}

\usepackage[capitalize]{cleveref}
\crefname{figure}{Fig.}{Figs.}
\crefname{table}{Tab.}{Tabs.}
\crefname{appendix}{App.}{Apps.}

\usepackage{todonotes}
\usepackage{soul}

\definecolor{TodoColor}{rgb}{1,0.7,0.6}

\usepackage{booktabs}   
\usepackage{colortbl}   
\usepackage[table]{xcolor} 
\usepackage[table,HTML]{xcolor}

\usepackage{enumitem}
\usepackage{placeins}
\newcommand{\huggingfacesmall}{\includegraphics[width=9px]{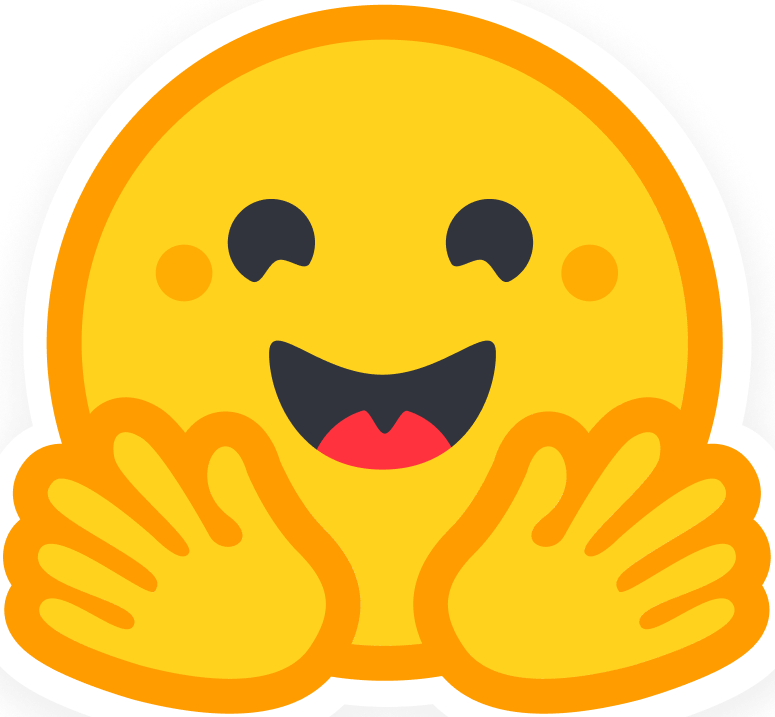}}

\usepackage{xstring}
\usepackage{seqsplit}

\newcommand{\hfmodel}[1]{%
    \StrBehind{#1}{/}[\hfmodelshortname]%
    \StrSubstitute{\hfmodelshortname}{-}{-\allowbreak}[\hfmodelbreakable]%
    \texttt{\hfmodelbreakable}\footnote{\huggingfacesmall{} \href{https://huggingface.co/#1}{#1}}%
}

\usepackage{amssymb}
\usepackage{pifont}

\usepackage{hyperref}

\usepackage{tcolorbox}
\tcbuselibrary{skins}
\newtcolorbox{promptbox}[1]{
    colback=gray!8,
    colframe=gray!45,
    title=#1,
    fonttitle=\small\bfseries,
    fontupper=\small\ttfamily,
    top=4pt, bottom=4pt
}

\usepackage{url}
\usepackage{xurl}
\usepackage{subcaption}

\def\langs#1#2{#1$\rightarrow$#2}
\def\enhe{\langs{En}{He}}
\def\ende{\langs{En}{De}}
\def\encs{\langs{En}{Cs}}
\def\csen{\langs{Cs}{En}}
\def\pCref#1{(\Cref{#1})}
\def\en{En}
\def\Aref#1{Appendix~\ref{#1}}

%% file: content.tex
\begin{abstract}
Errors in speech translations reduce trustworthiness of Speech Translation (ST) systems and can have serious consequences.
Yet currently there is no established methodology for evaluating confidence and quality estimation of speech translations.
To initiate progress in this direction, we propose Speech Translation Error Labelling (STEL).
We create an annotation protocol, a small authentic end-to-end evaluation dataset, and we analyse how existing text-only and speech-processing systems perform the STEL task.
Our results show that text-only XCOMET and multimodal LLM Qwen2.5-Omni are able to perform the STEL task in roughly half the precision of humans. We also find that direct speech processing is necessary for the STEL task, and that the current text-only and speech-processing systems are complementary in labelling translation-only vs.\ speech-processing errors in ST.
\end{abstract}

\section{Introduction}

Speech Translation (ST) has the potential to reduce language barriers in many everyday situations. To enable its widespread adoption, including in domains where misunderstanding can be critical, such as healthcare, business, or law, it is necessary to make its output more trustworthy. Speech Translation Quality Estimation (ST QE) could be used to highlight potentially incorrectly translated words, which would allow ST users to respond appropriately, e.g.\ ask for clarification of the highlighted output spans. It could also enable automatic system selection 
and post-editing both by humans and generative large language models.


While automatic QE systems
already exist for text machine translation \cite{lavie-etal-2025-findings,Zhao2024FromHF,guerreiro-etal-2024-xcomet}, there is an open question whether automatic ST QE should use ASR transcripts or whether they should process speech directly to prevent compounding of errors and information loss.

Moreover, ST is usually used in specific communication situations that differ from text-only communication \cite{fantinuoli-prandi-2021-towards} for which Error Span Annotations datasets exist \cite{kocmi-etal-2024-error,kocmi-etal-2024-findings,kocmi-etal-2025-findings}. 
Speech communication participants typically have a communication goal and a knowledge of background context that often make them less sensitive to subtle grammar flaws, and may prefer fluent translations with hesitations and false starts removed. We need a new evaluation methodology that takes these aspects into account. 
%
%

We therefore propose the Speech Translation Error Labelling (STEL) task and make the following contributions: (1) we design an annotation protocol focused on end-to-end long-form speech translation \pCref{sec:protocol}; (2) we create a test set for four language directions, which required 6.5 hours of annotation effort \pCref{sec:dataset}; (3) we demonstrate the feasibility and state of the art of STEL using existing automatic systems, both text-only and speech processing, and (4) we investigate using text-only vs.\ speech-processing methods for STEL \pCref{sec:automatic}.
We release the dataset and code publicly.\footnote{
\url{https://github.com/CSTR-Edinburgh/STEL}
}




\section{STEL Annotation Protocol}
\label{sec:protocol}

The first contribution of this work is an STEL annotation protocol, which builds upon the existing Error Span Annotation for text MT \cite{kocmi-etal-2024-error}, but adds the ST user-centric and communication-oriented perspective \cite{fantinuoli-prandi-2021-towards}. 

In our annotation protocol we assign four severities of errors to erroneous subsequences of translations, called \textit{spans}. \emph{Critical} errors are mistranslations that change the meaning of the sentence and cannot be corrected by the listener.
A mistranslation of a name or term that could be inferred by a knowledgeable listener is a \emph{minor} error. Subtle, e.g.\ grammar errors, that may only affect comfort but not meaning, are \emph{negligible} errors. 
\emph{Redundancies} are, for example, verbosely translated repetitions, false starts, filler words, etc. While redundancies may or may not be translation errors, they typically divert focus from the communicative goal. 



In addition to error span detection and categories, STEL also includes segment-level DA ranging from 0-100. It is an auxiliary task which may be helpful for training automatic STEL systems, similarly as in text MT ESA.



\section{STEL Dataset Creation}
\label{sec:dataset}

The second contribution of this work is a STEL dataset primarily intended for evaluation of automatic ST error labelling systems.

We selected four language directions that represent authentic use cases: 
Czech (Cs) to English (En), and English to Czech, German (De), and Hebrew (He).
We reused documents from existing ST test sets, namely the \langs{Cs}{En} Robothon debate from the ELITR Test Set \cite{ansari-etal-2021-sltev}, and one randomly selected talk from the ACL6060 test set \cite{salesky-etal-2023-evaluating} for \langs{En}{\{Cs, De, He\}}.

We selected ST candidate systems as representatives of state of the art from three typically used groups of models that are publicly available for self-hosting: (1) \textbf{cascades of ASR and LLM for MT} because they tend to be top-performing \cite{papi2025hearingtranslateeffectivenessspeech}, but may be large and costly;
(2) \textbf{an end-to-end speech-to-text model} that tends to be small and high-performing;
(3) \textbf{a simultaneous system} that processes incremental input in low latency, allowing real-time interaction. 
Since all these systems either include ASR component in the cascade, or use a multi-task models, we used their ASR transcripts for automatic labelling in \cref{sec:automatic}.
Details about the used ST systems are in \Aref{ap:st_details}.

We translated the recordings with ST systems, and applied mWERSegmenter~\citep{post-hoang-2025-effects} to segment and align the candidate translation to the sentence segmentation of the reference translation that is available in the reused test sets. Since Hebrew and Czech translations are not available in ACL6060, we used manual segmentation and alignment for Hebrew and machine translated source transcript for mWERSegmenter for Czech.
We segmented the recordings without omissions to allow the annotators to take into account the inter-sentence context. 

We collected STEL annotations in Pearmut \cite{zouhar2026pearmut}, which is a flexible web application for multi-lingual evaluation campaigns.
We asked three volunteers, native Czech, German, and Hebrew speakers, who are English L2 speakers, to provide STEL annotations.
We instructed them to consider typical audience, their background knowledge, and their communication goal.
In total, they worked 6.5 hours and annotated 329 sentence-like segments in 32 minutes of audio. 
\Cref{tab:annotated_stats} summarizes the size of the annotated data.
We present the exact instructions and an overview of the counts and categories of annotated spans in  \cref{app:data_collection} in \cref{fig:pearmut_instruct} and \Cref{tab:error-counts} respectively.

As a consistency check and to obtain an upper bound for automatic systems, we asked the same \encs{} and \ende{} annotators to annotate the same data again for the second time one month after their first annotation.
We observe relatively high agreement of 72\% for the \encs{} annotator, and 56 \% for the \ende{} annotator. 
We explain the gap from the ideal 100\% agreement by a combination of factors. First, by  ambiguity of natural language, which can not be avoided, but can be handled e.g.\ by using multiple independent annotations as references. The second group of factors include 
the low experience of annotators, their limited attention, and by annotation guidelines being not specific in corner cases. These factors can be improved in additional work, which is beyond the available capacity. Nevertheless, we conclude that our annotations are reasonable for demonstrating the systems evaluation and for providing evidence for our research questions.

\input{results-tables/annotated_stats}


\section{Automatic Labelling}
\label{sec:automatic}

Our third contribution is an experimental study of existing automatic systems that can perform STEL, aiming to answer: (1) How well is it possible to perform STEL with the current automatic systems? (2) Do the automatic STEL systems benefit from having access to the source audio?

\def\model{\textbf{~}}
\def\microfw{\textbf{F1 w}}
\def\microfuw{\textbf{F1 uw}}
\def\microp{Prec.}
\def\micror{Rec.}
\def\macrofw{ma F1w}
\def\macrofuw{ma F1uw}
\def\pearson{Pear.}
\def\spearman{Spear.}
\def\kendall{\textbf{Kend.}}
\def\nan{-}
\def\XCOMETn{XCOMET}
\def\XCOMET{XCOMET (ASR)}
\def\XCOMETg{XCOMET (gold)}
\def\Qwenn{Qwen}
\def\Qwen{Qwen (ASR+audio)}
\def\Qweng{Qwen (gold+audio)}
\def\secondannotation{2nd annotation}
\input{results-tables/stel_horizontal}

\subsection{Automatic STEL Systems}
\label{sec:stel-systems}

We evaluate representatives of two state-of-the-art systems that can perform STEL in many language directions including the ones in this study: a specialized text-MT QE model, and a general-purpose multimodal LLM.  Inference is performed on a single NVIDIA A100-SXM4-40GB GPU. Prompts and inference parameters are in \Aref{app:models}.

\paragraph{XCOMET}
\hfmodel{Unbabel/XCOMET-XL}~\citep{guerreiro-etal-2024-xcomet} is a
state-of-the-art reference-free QE model trained specifically to detect translation errors in text-MT.  Given a source sentence and a candidate translation, it produces a sentence-level quality score together with a list of error spans. Since we primarily focus on the real-life use case, where the gold source transcript is unavailable, we process XCOMET with an ASR transcript. The variant with gold transcript is only for contrastive analysis.

\paragraph{Qwen}
\hfmodel{Qwen/Qwen2.5-Omni-7B}~\citep{xu2025qwen25omnitechnicalreport} is a state-of-the-art
multimodal LLM capable of processing both text and audio in many languages.
We prompt it to produce a sentence-level quality score together with error
spans in the target text. We experimented with three parameters: modality, spans with/without severities of errors, and using the context of adjacent segments. However, we primarily report the variant with an ASR transcript and audio, with severities, and no adjacent context. 

Since neither XCOMET nor Qwen, unlike the ST systems, provide their own segmentation of the long-form source of multiple consecutive sentences into their individually processed units, we follow the common ST practice of using the ``gold'' source audio segmentation from the test sets. We leave the more realistic segmentation to future work.

\subsection{Meta-Evaluation}

We evaluate the STEL systems against human annotations along two dimensions:
span detection quality and score-level correlation.

\paragraph{Span detection.}
We measure how well predicted error spans overlap with human-annotated spans using character-level F1, following the WMT25 Error Span Annotation (ESA)
evaluation protocol~\citep{lavie-etal-2025-findings}. For each sample, we convert predicted and reference spans to sets of character
indices and 
we report \emph{unweighted micro F1} over their overlap. 
We also report the \emph{severity-weighted micro F1},
following \citet{lavie-etal-2025-findings}: same-severity overlaps receive full credit, while cross-severity overlaps receive half credit.
Since \mbox{XCOMET} predicts three (Critical, Minor, Major) severity labels, we treat a predicted \emph{Minor} as correct for gold \emph{Minor} or \emph{Negligible}, and a predicted \emph{Major} or \emph{Critical} as correct for gold \emph{Major} or \emph{Redundancy}.

\paragraph{Score correlation.}
We measure correlation to DA scores using Kendall's $\tau$ \citep{kendall}.

\subsection{Results}
\label{sec:results}

\Cref{tab:main-results} shows results of automatic systems using ASR, contrasted to the same systems using gold transcripts and the second human annotation. In general, we observe that \textbf{automatic systems are able to perform the STEL task with approximately half the quality of humans} (e.g. \encs{} F1 unweighted XCOMET 38.7 vs. human 71.9.). Both XCOMET and Qwen perform better in span identification than in assigning the severity labels (unweighted F1 scores are higher than weighted). XCOMET outperforms Qwen in three language directions, but not in \enhe{}, which may be caused by low amount of Hebrew training data in COMET, which 
leads to nearly no span predictions and unweighted F1 only 1.8, compared to 14.3 with gold transcript. The gold transcripts improve the error span labelling over the same model setups with ASR, probably due to higher quality over ASR, but only by little margin of 0 to 3.6, which suggests that most of the STEL task complexity is in other challenges than speech processing, for instance, translation modelling, not using inter-segment context, or segmentation to processing units.
Exceptions are \enhe{}, and \csen{} unweighted F1 for Qwen, but not weighted, which may be caused by lower quality of the \encs{} gold transcript. 

We also observe that the systems are able to provide segment-level DA with varying quality over language directions and models. Comparisons of Kendall's $\tau$ for Qwen on \ende{} vs.\ \encs{} (32.5 vs.\ 2.1) may indicate Qwen limitations in generalization, multitasking, or prompting, or the quality of DA annotations over annotators.

\def\werzero{wer=0}
\def\wernonzero{wer>0}
\begin{table}[]
    \centering
    \footnotesize
    \setlength{\tabcolsep}{5pt}
    \begin{tabular}{lcccc}
    \toprule
 &  \multicolumn{2}{c}{\microfw} & \multicolumn{2}{c}{\microfuw}  \\ 
\textbf{Avg. of all langs.} & \textbf{\wernonzero} & \textbf{\werzero} & \textbf{\wernonzero} & \textbf{\werzero}  \\ 
 \midrule
\XCOMETn{}$^{ASR}$ & 10.78 & \textbf{9.95} & 15.08 & \textbf{12.24 }\\ 
\Qwenn{}$^{ASR+audio}$ & 12.98 & 8.80 & \textbf{15.96} & 11.23 \\ 
\Qwenn{}$^{ASR}$ & 11.84 & 7.03 & 15.24 & 9.16 \\ 
\Qwenn{}$^{audio}$ & \textbf{13.11} & 7.04 & 15.15 & 9.00 \\ 
\bottomrule
    \end{tabular}
    \caption{STEL scores for speech-processing errors (\wernonzero) vs.\ translation-only errors (\werzero) for the realistic STEL systems.
    Maximum scores in bold.}
    \label{tab:wer-results}
\end{table}

\subsection{Speech vs. Text Translation Errors}
\label{sec:wer}

Next, we focus on the second research question, on speech vs.\ text-only processing in the automatic STEL systems. We presume two groups of ST errors by their origin: in ASR (or in the speech processing component of end-to-end ST) vs.\ translation-only errors. We therefore split the annotated error spans into these groups, depending whether the corresponding ASR spans (detected with AwesomeAlign, \citealp{dou-neubig-2021-word}, after tokenization with Moses \citealp{koehn-etal-2007-moses}) are correct (WER=0 when compared to the gold transcript), or include an error (WER>0). 
We observe that these two groups have nearly the same size. 


\Cref{tab:wer-results} summarizes the scores for both groups. We observe that translation-only errors are better detected by XCOMET, while speech-processing errors are better detected by Qwen using either audio (weighted F1), or ASR and audio (unweighted F1). We conclude that \textbf{speech-processing is a necessary component of the STEL methods,} and that the \textbf{current text-only and speech-processing STEL systems are complementary}.

\section{Recommendations}

In our experiments, we explored two STEL systems that used the internal ASR transcripts of the evaluated ST systems. While the text-based \XCOMETn{} is cheaper to run and excels at predicting translation-only errors, \Qwenn{} is more robust to ASR errors because it can leverage the speech input. Based on our experience, we would recommend choosing systems for future evaluations based on the language pairs under consideration, taking into account the amount of target-language text seen during training of \XCOMETn/\Qwenn, and the accuracy of the ASR system in the source language.

One approach to making STEL predictions more accurate is to reduce the gap to gold transcripts by improving the accuracy of the ASR systems used for STEL. This could be achieved by using a strong external ASR system, which would also allow black-box evaluation of ST systems that do not produce ASR transcripts. The external ASR system could also be further improved by using ASR post-correction.
We plan to evaluate using the strong external ASR system in future work.





The last criterion to consider is the severity taxonomy and how easily it can be customised. Since XCOMET cannot be easily adapted to predict new severity classes, it may be better to use speech LLMs, which can predict new severity labels with in-context learning.

\section{Conclusion}

In this paper, we introduced the Speech Translation Error Labelling task (STEL), which is crucial for making speech translation systems more trustworthy, especially in settings where errors can lead to critical consequences. We proposed the STEL annotation protocol for annotating spans of incorrect translations and assigning severity to these errors. We created and annotated a small evaluation dataset for four language directions, and we ana\-lysed the performance of two existing automatic systems on this task. Our results show that it is possible to label speech translation errors automatically, although current methods have limitations and achieve roughly half the performance of humans. We provide an evidence that future better performing methods may need to process speech directly, rather than relying on ASR, and that the current text-only and speech-processing systems are  complementary in labelling translation-only vs.\ speech-processing ST errors.

\section{Limitations}
One limitation of this work is the limited coverage of language directions, domains, and speech documents -- only two documents, 14 and 10 minutes long. Other domains and documents may include phenomena that are more challenging for speech processing, such as speaker characteristics (gender, age, accents, idiolect, dialect), emotion, etc.; or phenomena more challenging for translation, such as specific terminology, context, ambiguities, rare words, etc.

We warn against considering every annotated or omitted error span as ground truth without considering the limiting factors: the number and availability of annotators -- one annotator per language direction, 6.5 hours of annotation effors in total --- the limited experience in language annotations, and low variability of annotators. The annotated dataset is primarily intended for measuring progress in development of automatic STEL systems by aggregating scores across the whole dataset. 


There was also limited investigation of the optimal ST and STEL systems and their parameters.

\section{Ethical Considerations}

The annotators participated voluntarily. No personal data were collected nor published. The participating annotators were informed openly about their participations and planned used of their outcomes. 
No ethics board was involved, as no regulation of any institutions or country where this work was carried out requires it for this kind of work.

In this paper, AI assistants were used for assistance with the coding and formatting of tables. The authors reviewed their outputs.

All artifacts (datasets, models, software) were used in accordance with their licence and permitted use. The created artifact is equipped with limitations section and primarily intended use.


\section*{Acknowledgements}
This work has received funding from the European Union's Horizon research and innovation programme under grant agreement No 101135798, project Meetween (My Personal AI Mediator for Virtual MEETtings BetWEEN People), and by Czech Operational Program OP JAK, the MSCA CZ project MSCA Fellowships -- Charles University 4, CZ.02.01.01/00/22\_010/0013392, ``LCT''.
Ondrej Klejch was supported by the Scottish Government (Grant name: ``Ecosystem for Interactive Speech
Technologies'').

%% file: results-tables/annotated_stats.tex
\begin{table}[t]
\centering
\footnotesize
\begin{tabular}{l@{\hskip 0.5cm}rr@{\hskip 0.5cm}r}
\toprule
\textbf{Language directions} & \textbf{\# Seg.} & \textbf{Dur.} & \textbf{Work} [m.] \\
\midrule
\langs{Cs}{En} & 169 & 14:06 & 126 \\
\langs{En}{Cs} (twice) & 37  & 3:45  & 51 (+41) \\
\langs{En}{De} (twice) & 37  & 3:45  & 49 (+43) \\
\langs{En}{He} & 86  & 10:34 &  78 \\
\midrule
\textbf{Total:} & 329 & 32:10 & 388 \\
\bottomrule
\end{tabular}
\caption{Annotation statistics.
For \langs{En}{Cs} and \langs{En}{De}, we annotated the same data twice. For \langs{En}{He}, we have two ST candidates, for the others we have three.}
\label{tab:annotated_stats}
\end{table}

%% file: results-tables/stel_horizontal.tex
\begin{table*}[t]
    \centering
    \footnotesize
    \setlength{\tabcolsep}{3pt}
    \resizebox{1.0\linewidth}{!}
    {%
    \begin{tabular}{lc@{~}c@{~}cc@{~}c@{~}cc@{~}c@{~}cc@{~}c@{~}c} 
    \toprule
& \multicolumn{3}{c}{\textbf{\langs{Cs}{En}}} & \multicolumn{3}{c}{\textbf{\langs{En}{Cs}}} & \multicolumn{3}{c}{\textbf{\langs{En}{De}}} & \multicolumn{3}{c}{\textbf{\langs{En}{He}}} \\
\cmidrule(lr){2-4}\cmidrule(lr){5-7}\cmidrule(lr){8-10}\cmidrule(lr){11-13}
& \microfw  & \microfuw & \kendall & \microfw & \microfuw & \kendall & \microfw & \microfuw & \kendall & \microfw & \microfuw & \kendall \\
\midrule
\XCOMETn{}$^{ASR}$
    & $\textbf{18.4}_{+0.9}$ & $\textbf{22.3}_{+0.8}$  & $\textbf{23.5}_{-1.7}$
    & $\textbf{31.8}_{\phantom{-}0.0}$  & $\textbf{38.7}_{+0.4}$ & $11.5_{+8.7}$
    & $20.5_{+0.9}$ & $\textbf{30.5}_{-2.0}$  & $26.1_{+10.1}$
    & $1.0_{+9.2}$  & $1.8_{+12.5}$  & $1.5_{+22.3}$ \\
\Qwenn$^{ASR+audio}$
    & $14.3_{+0.6}$ & $18.1_{-0.4}$ & $12.6_{+4.5}$
    & $24.0_{+1.3}$ & $29.9_{+2.6}$ & $2.1_{+6.7}$
    & $\textbf{27.8}_{+0.2}$ & $30.1_{+0.4}$ & $32.5_{+20.7}$
    & $\textbf{13.2}_{-1.5}$ & $\textbf{17.5}_{-2.1}$ & $41.0_{-1.2}$ \\
\Qwenn$^{ASR}$
    & $15.0_{+1.2}$ & $17.5_{+0.4}$ & $13.3_{-5.3}$
    & $23.1_{-2.4}$ & $29.7_{-5.0}$ & $2.9_{-15.3}$
    & $21.4_{-2.6}$ & $24.2_{-2.0}$ & $25.0_{-27.3}$
    & $9.6_{-0.4}$ & $14.7_{+0.3}$ & $-3.0_{-44.7}$ \\
\Qwenn$^{audio}$
    & $11.8$\phantom{$_{-0.0}$} & $14.7$\phantom{$_{-0.0}$} & $3.5$\phantom{$_{-0.0}$}
    & $27.1$\phantom{$_{-0.0}$} & $30.4$\phantom{$_{-0.0}$} & $\textbf{13.2}$\phantom{$_{-0.0}$}
    & $23.4$\phantom{$_{-0.0}$} & $26.9$\phantom{$_{-0.0}$} & $43.0$\phantom{$_{-0.00}$}
    & $11.0$\phantom{$_{-0.0}$} & $12.8$\phantom{$_{-0.0}$} & $\textbf{41.6}$\phantom{$_{-0.0}$} \\

\midrule
\cellcolor{white}2nd annotation
    & \cellcolor{white}$-$ & \cellcolor{white}$-$ & \cellcolor{white}$-$
    & \cellcolor{white}72.8\phantom{$_{-0.0}$} & \cellcolor{white}71.9\phantom{$_{-0.0}$} & \cellcolor{white}51.4\phantom{$_{-0.0}$}
    & \cellcolor{white}52.5\phantom{$_{-0.0}$} & \cellcolor{white}55.5\phantom{$_{-0.0}$} & \cellcolor{white}63.1\phantom{$_{-0.0}$}
    & \cellcolor{white}$-$ & \cellcolor{white}$-$ & \cellcolor{white}$-$ \\

\bottomrule
    \end{tabular}%
    }
    \caption{Automatic STEL results. The meta-evaluation scores are micro weighted and unweighted F1 in \% (the higher, the better) and Kendall's $\tau$ multiplied by 100 (range -100 to 100, the higher, the better). 
    Subscripts indicate the difference when using gold transcripts.}
    \label{tab:main-results}
\end{table*}

%% file: appendix.tex
\section{Related Work}

Quality Estimation is a broad area of methods for simulating or facilitating costly human quality assessments. Existing work on ST QE focuses on ST metrics -- the cases when reference translation is available \cite{Cettolo2025HowTE,machacek-etal-2023-mt,post-hoang-2025-effects}, or on segment-level Direct Assessment \cite{han-etal-2024-speechqe,chen-etal-2023-blaser}, which is less informative than error span labelling, but possible to perform because there exist evaluation data. To our knowledge, no evaluation data exist for ST error span labelling.

\citet{kocmi-etal-2024-error} propose Error Span Annotation, a task in which human annotators highlight spans, erroneous subsequences of candidate translations, and assign them severity classes. They also add segment-level Direct Assessment (DA) quality scores. Error Span Annotation is proved to be easy to learn by annotators, and informative enough to evaluate candidate translation systems. It is widely used in text MT evaluation campaigns (WMT, \citealp{kocmi-etal-2024-findings,kocmi-etal-2025-findings}), and as a task for automatic systems \cite{zouhar-etal-2025-ai}. There are supervised neural models, such as XCOMET-XL \cite{guerreiro-etal-2024-xcomet} for text MT, that are designed for Error Span Annotation and DA. These tasks can also be performed by general-purpose Large Language Models, such as the multimodal Qwen2.5-Omni \cite{xu2025qwen25omnitechnicalreport}, but ST error span labelling with LLMs has not yet been studied.

However, Error Span Annotation is primarily designed for text MT. As \citet{fantinuoli-prandi-2021-towards} propose, ST is typically supposed to facilitate a specific communication goal of participants at a real communicative event. They propose a user-centric and communication-oriented ST evaluation, which we follow up in this work.

\section{ST Model Details}
\label{ap:st_details}

The selected models are the following:

\begin{enumerate}
    \item \textbf{Cascades of ASR and LLM for MT:} 
We select Whisper-large-v3 \cite{Whisper-paper}  as ASR and Gemma3:12b-it-qat as MT \cite{gemmateam2025gemma3technicalreport} for all directions except for \enhe{} where LLama3:70b \cite{grattafiori2024llama3herdmodels} scores better.

We used WhisperX (\url{https://github.com/m-bain/whisperX}).

We used Gemma3 on complete WhisperX segments that fit into 3000 characters. Llama3 processed the whole document in one step.

\item \textbf{End-to-end speech-to-text:} 
Canary-v2-1B \cite{sekoyan2025canarybv} for all language directions except for \enhe{}, for which there is no publicly available end-to-end ST.
We used the script for chunked inference: \url{https://github.com/NVIDIA-NeMo/NeMo/blob/main/examples/asr/asr_chunked_inference/aed/speech_to_text_aed_chunked_infer.py}.

\item \textbf{Simultaneous systems:} 
We use SimulStreaming \url{https://github.com/ufal/SimulStreaming} \cite{machacek-polak-2025-simultaneous} 
that integrates end-to-end speech-to-text Whisper-large-v3 for \csen{}, and Whisper-large-v3 as \en{} ASR and Tower+-9B \cite{rei2025tower} for \encs{} and \ende{}, and Gemma2-9B-it for \enhe{}.

\end{enumerate}

We selected the ST systems as representatives of state of the art for given language directions from the three typically used groups of models that are publicly available for self-hosting on an affordable hardware, where possible. 
We therefore selected models that have maximum 12B parameters and can be deployed on one NVIDIA A40 GPU, with an exception of Llama3:70b, which we used via a self-hosted OpenWebUI service that used 80GB H100 GPU in the backend.





\section{Dataset and Human Annotation Details}\label{app:data_collection}

\begin{figure*}
    \centering
    \includegraphics[width=1.0\linewidth]{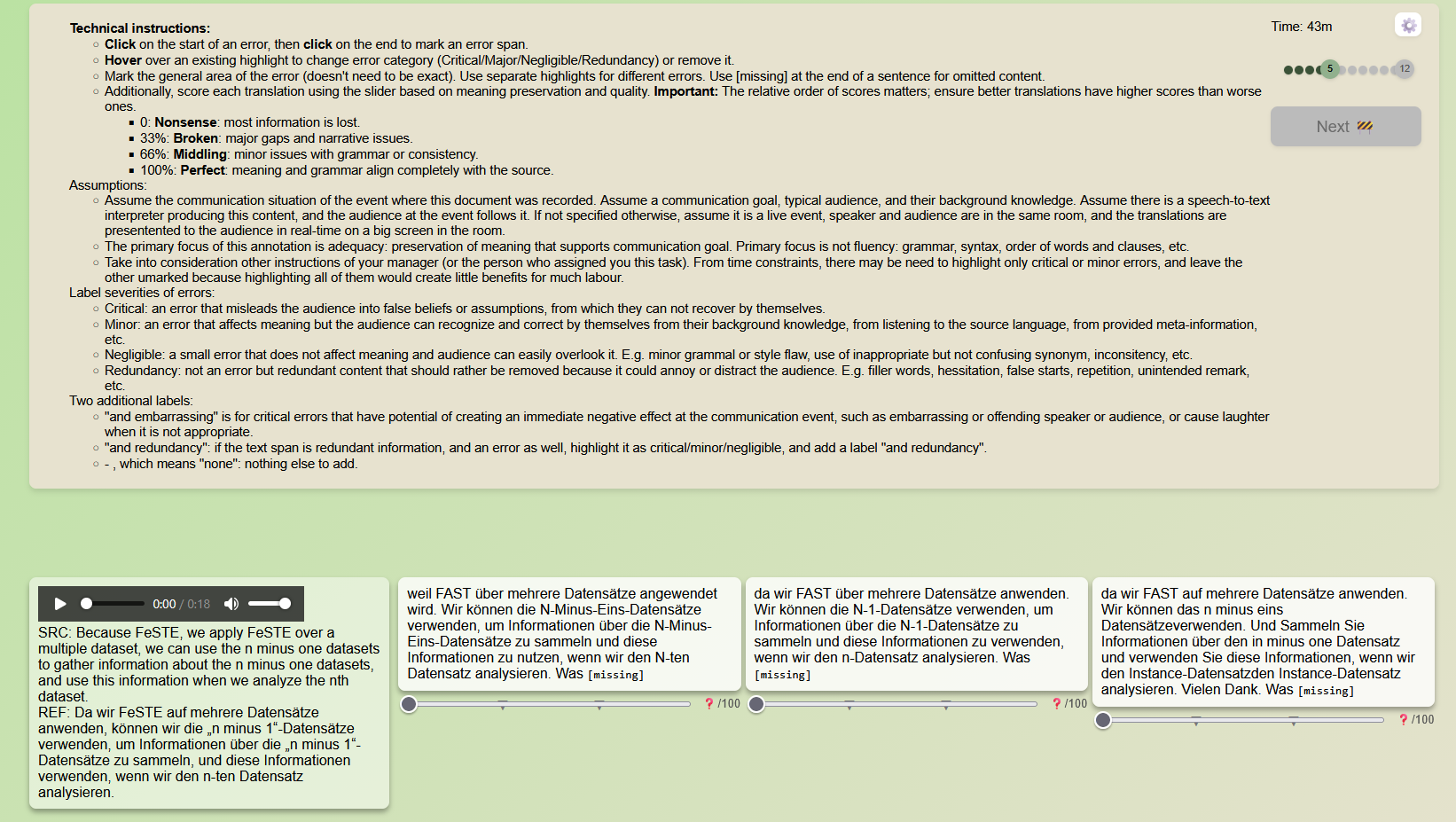}
    \caption{Screenshot of the annotation platform with instructions for the annotators. It is a version of Pearmut \citep{zouhar2026pearmut} that we extended with our STEL protocol.}
    \label{fig:pearmut_instruct}
\end{figure*}

See \Cref{tab:error-counts} and \Cref{fig:pearmut_instruct}.

\input{results-tables/error_counts}

\section{STEL Model Details}
\label{app:models}

\input{figures/prompts}

\paragraph{XCOMET-XL.}
\hfmodel{Unbabel/XCOMET-XL} is run with batch size~8 on a single NVIDIA A100-SXM4-40GB GPU.

\paragraph{Qwen2.5-Omni.}
\hfmodel{Qwen/Qwen2.5-Omni-7B}  uses greedy decoding and \texttt{max\_new\_tokens=512}. Each call consists of a system prompt and a modality-specific user message. The prompts are listed in \cref{fig:prompts-context} for prompts with context and \cref{fig:prompts-no-context}.

\subsection{Qwen Parameter Selection}

For inspecting results in \Cref{sec:results}, we selected the Qwen parameters in the following way:
(1) we excluded the less realistic setups with gold transcripts; (2) with severities of errors, because it is our primary task. Secondary task is error detection without classifying the severities. (3) We averaged the scores for all language directions, and selected the micro F1 weighted top performing variant of source modality and context length.

\Cref{tab:qwen-selection} shows the scores. The selected setup is Qwen with ASR and audio, and with no adjacent context.

\input{results-tables/selectqwen}

\section{Detailed Results}

\Cref{tab:qwen-avg-all,tab:qwen-avg-kendall} summarize scores of all inspected model variants, sorted by \microfuw{} or by Kendall's $\tau$.

\input{results-tables/avg_all_models}

\input{results-tables/avg_all_kendall}

\section{Word Alignment}

For the ASR to ST candidate alignment in \Cref{sec:wer}, 
we used AwesomeAlign (\url{https://github.com/neulab/awesome-align}) with the model \texttt{bert-base-multilingual-cased}.

For tokenization, we used \texttt{tokenizer.perl}, version 1.1, from Moses \cite{koehn-etal-2007-moses}, with the known language ID parameter for the processed language.

%% file: results-tables/error_counts.tex
\begin{table*}[t]
    \centering
    \setlength{\tabcolsep}{4pt}
    \footnotesize
\begin{tabular}{lccccc@{\hskip 0.6cm}ccccc@{\hskip 0.6cm}ccccc@{\hskip 0.6cm}ccccc}
    \toprule
    & \multicolumn{5}{c@{\hskip 0.6cm}}{\textbf{\csen}} & \multicolumn{5}{c@{\hskip 1cm}}{\textbf{\encs}} & \multicolumn{5}{c@{\hskip 0.6cm}}{\textbf{\ende}} & \multicolumn{5}{c@{\hskip 0.6cm}}{\textbf{\enhe}} \\
    \textbf{system} & \textbf{C} & \textbf{M} & \textbf{N} & \textbf{R} & \textbf{T} & \textbf{C} & \textbf{M} & \textbf{N} & \textbf{R} & \textbf{T} & \textbf{C} & \textbf{M} & \textbf{N} & \textbf{R} & \textbf{T} & \textbf{C} & \textbf{M} & \textbf{N} & \textbf{R} & \textbf{T} \\
    \midrule
    cascade      &  8 & 24 & 18 & 20 &  70 & 20 & 18 &  5 & 2 & 45 & 12 &  9 & 0 & 1 & 22 & 2 & 14 & 11 &  0 & 27 \\
    end2end      & 34 & 51 & 25 & 19 & 129 & 24 & 27 &  3 & 1 & 55 & 17 & 14 & 2 & 5 & 38 & - &  - &  - &  - &  -  \\
    simultaneous & 42 & 69 & 20 & 18 & 149 & 25 & 23 & 13 & 6 & 67 & 28 & 14 & 3 & 5 & 50 & 8 & 40 & 48 & 18 & 114  \\
    \bottomrule
\end{tabular}
    \caption{Counts of annotated error spans and categories (Critical, Minor, Negligible, Redundancy, Total).}
    \label{tab:error-counts}

\end{table*}

%% file: figures/prompts.tex
\begin{figure*}[t]
\begin{promptbox}{With severity --- no context}
\textbf{System:} You are a machine translation quality estimator.
Given \underline{\textnormal{\textit{source description}}} and its translation, you must:\\[2pt]
1.~Assign an overall quality score between 0.0 (completely wrong) and 1.0 (perfect).\\
2.~Identify error spans in the translation text with their severity.\\[4pt]
Respond ONLY with a valid JSON object in this exact format:\\
\texttt{\{"score": <float 0.0--1.0>, "error\_spans": [\{"text": "<exact substring>",
"severity": "<minor|major|neutral|redundancy>"\}, ...]\}}\\
If there are no errors, use an empty list for \texttt{error\_spans}. Output nothing else.\\[6pt]
\textbf{User (text):}~[Target --- evaluate this]\\
Source:~\underline{\textnormal{\textit{source sentence}}} \quad Translation:~\underline{\textnormal{\textit{MT output}}}\\[4pt]
\textbf{User (audio):}~[Target --- evaluate this]\\
\textlangle audio\textrangle \quad Translation:~\underline{\textnormal{\textit{MT output}}}\\[4pt]
\textbf{User (text+audio):}~[Target --- evaluate this]\\
\textlangle audio\textrangle\\
Transcript:~\underline{\textnormal{\textit{source sentence}}} \quad Translation:~\underline{\textnormal{\textit{MT output}}}
\end{promptbox}

\begin{promptbox}{Without severity --- no context}
\textbf{System:} You are a machine translation quality estimator.
Given \underline{\textnormal{\textit{source description}}} and its translation, you must:\\[2pt]
1.~Assign an overall quality score between 0.0 (completely wrong) and 1.0 (perfect).\\
2.~Identify error spans in the translation text.\\[4pt]
Respond ONLY with a valid JSON object in this exact format:\\
\texttt{\{"score": <float 0.0--1.0>, "error\_spans": [\{"text": "<exact substring>"\}, ...]\}}\\
If there are no errors, use an empty list for \texttt{error\_spans}. Output nothing else.\\[6pt]
\textbf{User (text):}~[Target --- evaluate this]\\
Source:~\underline{\textnormal{\textit{source sentence}}} \quad Translation:~\underline{\textnormal{\textit{MT output}}}\\[4pt]
\textbf{User (audio):}~[Target --- evaluate this]\\
\textlangle audio\textrangle \quad Translation:~\underline{\textnormal{\textit{MT output}}}\\[4pt]
\textbf{User (text+audio):}~[Target --- evaluate this]\\
\textlangle audio\textrangle\\
Transcript:~\underline{\textnormal{\textit{source sentence}}} \quad Translation:~\underline{\textnormal{\textit{MT output}}}
\end{promptbox}
\caption{Prompts for Qwen2.5-Omni-7B without context. \underline{\textnormal{\textit{source description}}} is instantiated as \emph{a source sentence} (text), \emph{a source speech audio} (audio), or \emph{a source speech audio with its transcript} (text+audio).}
\label{fig:prompts-no-context}
\end{figure*}

\begin{figure*}[t]
\begin{promptbox}{With severity --- with context ($n$ sentences before/after)}
\textbf{System:} You are a machine translation quality estimator.
Given surrounding context sentences and a target \underline{\textnormal{\textit{source description}}}
with its translation, you must:\\[2pt]
1.~Assign an overall quality score between 0.0 (completely wrong) and 1.0 (perfect).\\
2.~Evaluate ONLY the target translation (marked [Target]).
Identify error spans in the translation text with their severity.\\[4pt]
Respond ONLY with a valid JSON object in this exact format:\\
\texttt{\{"score": <float 0.0--1.0>, "error\_spans": [\{"text": "<exact substring>",
"severity": "<minor|major|neutral|redundancy>"\}, ...]\}}\\
If there are no errors, use an empty list for \texttt{error\_spans}. Output nothing else.\\[6pt]
\textbf{User (text):}~[Context before]\\
Source:~\underline{\textnormal{\textit{ctx src$_1$}}} \quad Translation:~\underline{\textnormal{\textit{ctx MT$_1$}}} \quad $\cdots$\\
{[Target --- evaluate this]}\\
Source:~\underline{\textnormal{\textit{source sentence}}} \quad Translation:~\underline{\textnormal{\textit{MT output}}}\\
{[Context after]}\\
Source:~\underline{\textnormal{\textit{ctx src$_1$}}} \quad Translation:~\underline{\textnormal{\textit{ctx MT$_1$}}} \quad $\cdots$\\[4pt]
\textbf{User (audio):}~[Context before]\\
\textlangle audio\textrangle \quad Translation:~\underline{\textnormal{\textit{ctx MT$_1$}}} \quad $\cdots$\\
{[Target --- evaluate this]}\\
\textlangle audio\textrangle \quad Translation:~\underline{\textnormal{\textit{MT output}}}\\
{[Context after]}\\
\textlangle audio\textrangle \quad Translation:~\underline{\textnormal{\textit{ctx MT$_1$}}} \quad $\cdots$\\[4pt]
\textbf{User (text+audio):}~[Context before]\\
\textlangle audio\textrangle \quad Transcript:~\underline{\textnormal{\textit{ctx src$_1$}}} \quad Translation:~\underline{\textnormal{\textit{ctx MT$_1$}}} \quad $\cdots$\\
{[Target --- evaluate this]}\\
\textlangle audio\textrangle\\
Transcript:~\underline{\textnormal{\textit{source sentence}}} \quad Translation:~\underline{\textnormal{\textit{MT output}}}\\
{[Context after]}\\
\textlangle audio\textrangle \quad Transcript:~\underline{\textnormal{\textit{ctx src$_1$}}} \quad Translation:~\underline{\textnormal{\textit{ctx MT$_1$}}} \quad $\cdots$
\end{promptbox}

\begin{promptbox}{Without severity --- with context ($n$ sentences before/after)}
\textbf{System:} You are a machine translation quality estimator.
Given surrounding context sentences and a target \underline{\textnormal{\textit{source description}}}
with its translation, you must:\\[2pt]
1.~Assign an overall quality score between 0.0 (completely wrong) and 1.0 (perfect).\\
2.~Evaluate ONLY the target translation (marked [Target]).
Identify error spans in the translation text.\\[4pt]
Respond ONLY with a valid JSON object in this exact format:\\
\texttt{\{"score": <float 0.0--1.0>, "error\_spans": [\{"text": "<exact substring>"\}, ...]\}}\\
If there are no errors, use an empty list for \texttt{error\_spans}. Output nothing else.\\[6pt]
\textnormal{\textit{User messages identical to the with-severity variant above.}}
\end{promptbox}
\caption{Prompts for Qwen2.5-Omni-7B with context. We evaluate
context sizes $n \in \{1, 2, 5\}$; the template is the same for all sizes. \underline{\textnormal{\textit{source description}}} is instantiated as \emph{a source sentence} (text), \emph{a source speech audio} (audio), or \emph{a source speech audio with its transcript} (text+audio).}
\label{fig:prompts-context}
\end{figure*}

%% file: results-tables/selectqwen.tex
\begin{table*}[]
    \centering
    \footnotesize
    \begin{tabular}{lrrrrrrrrr}
    \toprule
    Avg. of all langs. & \microfw & \microfuw & \microp & \micror & \macrofw & \macrofuw & \pearson & \spearman & \kendall \\
    \midrule
Qwen (ASR+audio) & 19.8 & 23.9 & 16.6 & 50.9 & 17.8 & 37.6 & 27.3 & 25.9 & 22.0 \\ 
Qwen (ASR) & 17.3 & 21.5 & 14.8 & 50.5 & 16.1 & 32.8 & 8.9 & 11.6 & 9.6 \\ 
Qwen (ASR, ctx=1) & 17.4 & 21.5 & 14.1 & 62.6 & 16.4 & 30.9 & 12.1 & 10.6 & 9.0 \\ 
Qwen (audio) & 18.3 & 21.2 & 14.8 & 48.6 & 19.2 & 37.5 & 37.4 & 29.9 & 25.3 \\ 
Qwen (ASR, ctx=2) & 15.7 & 20.4 & 13.5 & 54.9 & 15.4 & 32.0 & 17.8 & 18.4 & 15.9 \\ 
Qwen (audio, ctx=2) & 18.0 & 18.2 & 14.9 & 41.1 & 16.2 & 34.1 & 14.5 & 15.7 & 13.2 \\ 
Qwen (ASR, ctx=5) & 15.2 & 18.2 & 12.7 & 42.6 & 15.4 & 31.9 & 6.8 & 11.1 & 9.4 \\ 
Qwen (audio, ctx=1) & 17.1 & 18.0 & 13.7 & 49.4 & 16.9 & 31.0 & 18.7 & 12.6 & 10.7 \\ 
Qwen (ASR+audio, ctx=1) & 16.2 & 17.0 & 13.9 & 30.6 & 15.8 & 37.6 & 11.8 & 11.9 & 10.3 \\ 
Qwen (ASR+audio, ctx=2) & 16.4 & 16.8 & 15.2 & 23.0 & 15.6 & 39.9 & 8.7 & 10.5 & 9.2 \\ 
Qwen (ASR+audio, ctx=5) & 14.5 & 13.9 & 13.2 & 16.4 & 13.8 & 41.9 & -5.5 & -7.5 & -6.5 \\ 
Qwen (audio, ctx=5) & 14.3 & 12.3 & 12.2 & 26.7 & 13.2 & 34.2 & 0.9 & -0.3 & -0.1 \\
    \bottomrule
    \end{tabular}
    \caption{Qwen results, average of all language directions, from which we selected the primary variant. Scores are averaged across all the ST systems, and sorted in decreasing order by micro F1 weighted. Scores description: \microfw{} = micro F1 weighted; \microfuw{} = micro F1 unweighted; Precision; Recall; macro F1 weighted/unweighted; Pearson, Spearman, Kendall. Left-most column shows the parameters, ctx is the number of adjacent segments as context.
}
    \label{tab:qwen-selection}
\end{table*}

%% file: results-tables/avg_all_models.tex
\begin{table*}[]
    \centering
    \footnotesize
    \begin{tabular}{lrrrrrrrrrr}
    \toprule
    Avg. of all langs. & \microfw & \microfuw & \microp & \micror & \macrofw & \macrofuw & \pearson & \spearman & \kendall \\
    \midrule 
XCOMET (gold) & 20.7 & 26.2 & 18.8 & 57.2 & 16.0 & 27.2 & 27.8 & 34.9 & 25.5 \\ 
Qwen (ASR+audio, no sev) & \nan & 24.7 & 17.3 & 49.3 & \nan & 39.6 & 29.0 & 27.5 & 23.4 \\ 
Qwen (gold, no sev) & \nan & 24.4 & 16.6 & 55.9 & \nan & 39.6 & 43.6 & 40.4 & 33.3 \\ 
Qwen (gold+audio) & 20.0 & 24.0 & 16.0 & 59.0 & 19.7 & 36.9 & 43.0 & 35.2 & 29.7 \\ 
Qwen (ASR+audio) & 19.8 & 23.9 & 16.6 & 50.9 & 17.8 & 37.6 & 27.3 & 25.9 & 22.0 \\ 
Qwen (gold, ctx=2, no sev) & \nan & 23.9 & 16.0 & 56.5 & \nan & 38.9 & 46.2 & 40.2 & 34.0 \\ 
XCOMET (ASR) & 17.9 & 23.3 & 17.3 & 40.8 & 13.1 & 33.5 & 18.1 & 21.3 & 15.7 \\ 
Qwen (gold) & 18.3 & 23.1 & 15.4 & 56.6 & 17.9 & 35.1 & 44.1 & 39.6 & 32.7 \\ 
Qwen (gold, ctx=1) & 18.3 & 23.1 & 15.0 & 66.3 & 17.9 & 32.5 & 44.0 & 33.8 & 28.3 \\ 
Qwen (gold, ctx=1, no sev) & \nan & 22.7 & 15.2 & 60.3 & \nan & 34.2 & 44.9 & 37.8 & 31.7 \\ 
Qwen (ASR, ctx=1, no sev) & \nan & 22.2 & 15.0 & 57.5 & \nan & 32.4 & 13.4 & 13.2 & 11.2 \\ 
Qwen (gold, ctx=2) & 18.5 & 22.1 & 14.5 & 58.4 & 18.4 & 34.2 & 46.7 & 40.3 & 34.4 \\ 
Qwen (gold+audio, no sev) & \nan & 22.1 & 15.0 & 52.1 & \nan & 37.7 & 44.5 & 37.1 & 31.4 \\ 
Qwen (ASR, ctx=2, no sev) & \nan & 22.0 & 15.2 & 51.8 & \nan & 34.4 & 14.8 & 15.1 & 12.9 \\ 
Qwen (gold, ctx=5, no sev) & \nan & 21.7 & 15.1 & 45.9 & \nan & 37.8 & 40.2 & 34.0 & 29.0 \\ 
Qwen (ASR) & 17.3 & 21.5 & 14.8 & 50.5 & 16.1 & 32.8 & 8.9 & 11.6 & 9.6 \\ 
Qwen (ASR, ctx=1) & 17.4 & 21.5 & 14.1 & 62.6 & 16.4 & 30.9 & 12.1 & 10.6 & 9.0 \\ 
Qwen (audio) & 18.3 & 21.2 & 14.8 & 48.6 & 19.2 & 37.5 & 37.4 & 29.9 & 25.3 \\ 
Qwen (ASR, ctx=2) & 15.7 & 20.4 & 13.5 & 54.9 & 15.4 & 32.0 & 17.8 & 18.4 & 15.9 \\ 
Qwen (gold, ctx=5) & 16.8 & 20.4 & 13.8 & 50.5 & 16.2 & 34.9 & 39.8 & 32.9 & 28.0 \\ 
Qwen (audio, no sev) & \nan & 20.3 & 15.2 & 42.3 & \nan & 38.4 & 38.5 & 30.8 & 25.8 \\ 
Qwen (ASR, no sev) & \nan & 20.0 & 14.1 & 44.6 & \nan & 32.9 & 8.2 & 10.8 & 8.5 \\ 
Qwen (ASR, ctx=5, no sev) & \nan & 19.3 & 14.0 & 40.0 & \nan & 33.2 & 8.0 & 9.6 & 8.2 \\ 
Qwen (gold+audio, ctx=2) & 19.0 & 19.0 & 16.2 & 29.0 & 18.7 & 39.0 & 16.1 & 19.1 & 16.2 \\ 
Qwen (audio, ctx=2) & 18.0 & 18.2 & 14.9 & 41.1 & 16.2 & 34.1 & 14.5 & 15.7 & 13.2 \\ 
Qwen (ASR, ctx=5) & 15.2 & 18.2 & 12.7 & 42.6 & 15.4 & 31.9 & 6.8 & 11.1 & 9.4 \\ 
Qwen (audio, ctx=1) & 17.1 & 18.0 & 13.7 & 49.4 & 16.9 & 31.0 & 18.7 & 12.6 & 10.7 \\ 
Qwen (gold+audio, ctx=1, no sev) & \nan & 17.9 & 14.2 & 35.7 & \nan & 37.1 & 26.8 & 16.3 & 13.8 \\ 
Qwen (gold+audio, ctx=1) & 17.4 & 17.8 & 14.4 & 36.3 & 17.6 & 36.8 & 25.0 & 14.9 & 12.5 \\ 
Qwen (ASR+audio, ctx=1, no sev) & \nan & 17.5 & 14.0 & 28.8 & \nan & 39.3 & 17.8 & 13.7 & 11.8 \\ 
Qwen (audio, ctx=2, no sev) & \nan & 17.1 & 14.8 & 34.0 & \nan & 35.2 & 12.7 & 14.8 & 12.5 \\ 
Qwen (ASR+audio, ctx=1) & 16.2 & 17.0 & 13.9 & 30.6 & 15.8 & 37.6 & 11.8 & 11.9 & 10.3 \\ 
Qwen (audio, ctx=1, no sev) & \nan & 16.9 & 12.9 & 41.8 & \nan & 32.1 & 24.6 & 19.9 & 17.2 \\ 
Qwen (ASR+audio, ctx=2) & 16.4 & 16.8 & 15.2 & 23.0 & 15.6 & 39.9 & 8.7 & 10.5 & 9.2 \\ 
Qwen (gold+audio, ctx=2, no sev) & \nan & 16.3 & 14.4 & 24.0 & \nan & 38.6 & 18.9 & 20.8 & 17.8 \\ 
Qwen (ASR+audio, ctx=5, no sev) & \nan & 14.9 & 17.0 & 17.9 & \nan & 40.9 & -7.5 & -5.8 & -4.9 \\ 
Qwen (gold+audio, ctx=5) & 14.3 & 14.7 & 13.3 & 20.3 & 16.0 & 42.8 & 3.8 & 4.5 & 4.2 \\ 
Qwen (gold+audio, ctx=5, no sev) & \nan & 14.0 & 14.5 & 19.6 & \nan & 41.1 & 0.7 & 1.7 & 1.8 \\ 
Qwen (ASR+audio, ctx=5) & 14.5 & 13.9 & 13.2 & 16.4 & 13.8 & 41.9 & -5.5 & -7.5 & -6.5 \\ 
Qwen (ASR+audio, ctx=2, no sev) & \nan & 13.4 & 13.5 & 17.2 & \nan & 38.6 & 8.8 & 11.0 & 9.5 \\ 
Qwen (audio, ctx=5, no sev) & \nan & 13.3 & 11.9 & 24.4 & \nan & 37.1 & -1.7 & -5.6 & -4.9 \\ 
Qwen (audio, ctx=5) & 14.3 & 12.3 & 12.2 & 26.7 & 13.2 & 34.2 & 0.9 & -0.3 & -0.1 \\
    \bottomrule
    \end{tabular}
    \caption{Scores of all inspected XCOMET and Qwen variants. Averaged across all language directions, sorted by \microfuw{} in descending order. Description of columns as in \Cref{tab:qwen-selection}.}
    \label{tab:qwen-avg-all}
\end{table*}

%% file: results-tables/avg_all_kendall.tex
\begin{table*}[]
    \centering
    \footnotesize
    \begin{tabular}{lrrrrrrrrrr}
    \toprule
    Avg. of all langs. & \microfw & \microfuw & \microp & \micror & \macrofw & \macrofuw & \pearson & \spearman & \kendall \\
    \midrule 
Qwen (gold, ctx=2) & 18.5 & 22.1 & 14.5 & 58.4 & 18.4 & 34.2 & 46.7 & 40.3 & 34.4 \\ 
Qwen (gold, ctx=2, no sev) & \nan & 23.9 & 16.0 & 56.5 & \nan & 38.9 & 46.2 & 40.2 & 34.0 \\ 
Qwen (gold, no sev) & \nan & 24.4 & 16.6 & 55.9 & \nan & 39.6 & 43.6 & 40.4 & 33.3 \\ 
Qwen (gold) & 18.3 & 23.1 & 15.4 & 56.6 & 17.9 & 35.1 & 44.1 & 39.6 & 32.7 \\ 
Qwen (gold, ctx=1, no sev) & \nan & 22.7 & 15.2 & 60.3 & \nan & 34.2 & 44.9 & 37.8 & 31.7 \\ 
Qwen (gold+audio, no sev) & \nan & 22.1 & 15.0 & 52.1 & \nan & 37.7 & 44.5 & 37.1 & 31.4 \\ 
Qwen (gold+audio) & 20.0 & 24.0 & 16.0 & 59.0 & 19.7 & 36.9 & 43.0 & 35.2 & 29.7 \\ 
Qwen (gold, ctx=5, no sev) & \nan & 21.7 & 15.1 & 45.9 & \nan & 37.8 & 40.2 & 34.0 & 29.0 \\ 
Qwen (gold, ctx=1) & 18.3 & 23.1 & 15.0 & 66.3 & 17.9 & 32.5 & 44.0 & 33.8 & 28.3 \\ 
Qwen (gold, ctx=5) & 16.8 & 20.4 & 13.8 & 50.5 & 16.2 & 34.9 & 39.8 & 32.9 & 28.0 \\ 
Qwen (audio, no sev) & \nan & 20.3 & 15.2 & 42.3 & \nan & 38.4 & 38.5 & 30.8 & 25.8 \\ 
XCOMET (gold) & 20.7 & 26.2 & 18.8 & 57.2 & 16.0 & 27.2 & 27.8 & 34.9 & 25.5 \\ 
Qwen (audio) & 18.3 & 21.2 & 14.8 & 48.6 & 19.2 & 37.5 & 37.4 & 29.9 & 25.3 \\ 
Qwen (ASR+audio, no sev) & \nan & 24.7 & 17.3 & 49.3 & \nan & 39.6 & 29.0 & 27.5 & 23.4 \\ 
Qwen (ASR+audio) & 19.8 & 23.9 & 16.6 & 50.9 & 17.8 & 37.6 & 27.3 & 25.9 & 22.0 \\ 
Qwen (gold+audio, ctx=2, no sev) & \nan & 16.3 & 14.4 & 24.0 & \nan & 38.6 & 18.9 & 20.8 & 17.8 \\ 
Qwen (audio, ctx=1, no sev) & \nan & 16.9 & 12.9 & 41.8 & \nan & 32.1 & 24.6 & 19.9 & 17.2 \\ 
Qwen (gold+audio, ctx=2) & 19.0 & 19.0 & 16.2 & 29.0 & 18.7 & 39.0 & 16.1 & 19.1 & 16.2 \\ 
Qwen (ASR, ctx=2) & 15.7 & 20.4 & 13.5 & 54.9 & 15.4 & 32.0 & 17.8 & 18.4 & 15.9 \\ 
XCOMET (ASR) & 17.9 & 23.3 & 17.3 & 40.8 & 13.1 & 33.5 & 18.1 & 21.3 & 15.7 \\ 
Qwen (gold+audio, ctx=1, no sev) & \nan & 17.9 & 14.2 & 35.7 & \nan & 37.1 & 26.8 & 16.3 & 13.8 \\ 
Qwen (audio, ctx=2) & 18.0 & 18.2 & 14.9 & 41.1 & 16.2 & 34.1 & 14.5 & 15.7 & 13.2 \\ 
Qwen (ASR, ctx=2, no sev) & \nan & 22.0 & 15.2 & 51.8 & \nan & 34.4 & 14.8 & 15.1 & 12.9 \\ 
Qwen (audio, ctx=2, no sev) & \nan & 17.1 & 14.8 & 34.0 & \nan & 35.2 & 12.7 & 14.8 & 12.5 \\ 
Qwen (gold+audio, ctx=1) & 17.4 & 17.8 & 14.4 & 36.3 & 17.6 & 36.8 & 25.0 & 14.9 & 12.5 \\ 
Qwen (ASR+audio, ctx=1, no sev) & \nan & 17.5 & 14.0 & 28.8 & \nan & 39.3 & 17.8 & 13.7 & 11.8 \\ 
Qwen (ASR, ctx=1, no sev) & \nan & 22.2 & 15.0 & 57.5 & \nan & 32.4 & 13.4 & 13.2 & 11.2 \\ 
Qwen (audio, ctx=1) & 17.1 & 18.0 & 13.7 & 49.4 & 16.9 & 31.0 & 18.7 & 12.6 & 10.7 \\ 
Qwen (ASR+audio, ctx=1) & 16.2 & 17.0 & 13.9 & 30.6 & 15.8 & 37.6 & 11.8 & 11.9 & 10.3 \\ 
Qwen (ASR) & 17.3 & 21.5 & 14.8 & 50.5 & 16.1 & 32.8 & 8.9 & 11.6 & 9.6 \\ 
Qwen (ASR+audio, ctx=2, no sev) & \nan & 13.4 & 13.5 & 17.2 & \nan & 38.6 & 8.8 & 11.0 & 9.5 \\ 
Qwen (ASR, ctx=5) & 15.2 & 18.2 & 12.7 & 42.6 & 15.4 & 31.9 & 6.8 & 11.1 & 9.4 \\ 
Qwen (ASR+audio, ctx=2) & 16.4 & 16.8 & 15.2 & 23.0 & 15.6 & 39.9 & 8.7 & 10.5 & 9.2 \\ 
Qwen (ASR, ctx=1) & 17.4 & 21.5 & 14.1 & 62.6 & 16.4 & 30.9 & 12.1 & 10.6 & 9.0 \\ 
Qwen (ASR, no sev) & \nan & 20.0 & 14.1 & 44.6 & \nan & 32.9 & 8.2 & 10.8 & 8.5 \\ 
Qwen (ASR, ctx=5, no sev) & \nan & 19.3 & 14.0 & 40.0 & \nan & 33.2 & 8.0 & 9.6 & 8.2 \\ 
Qwen (gold+audio, ctx=5) & 14.3 & 14.7 & 13.3 & 20.3 & 16.0 & 42.8 & 3.8 & 4.5 & 4.2 \\ 
Qwen (gold+audio, ctx=5, no sev) & \nan & 14.0 & 14.5 & 19.6 & \nan & 41.1 & 0.7 & 1.7 & 1.8 \\ 
Qwen (audio, ctx=5) & 14.3 & 12.3 & 12.2 & 26.7 & 13.2 & 34.2 & 0.9 & -0.3 & -0.1 \\ 
Qwen (audio, ctx=5, no sev) & \nan & 13.3 & 11.9 & 24.4 & \nan & 37.1 & -1.7 & -5.6 & -4.9 \\ 
Qwen (ASR+audio, ctx=5, no sev) & \nan & 14.9 & 17.0 & 17.9 & \nan & 40.9 & -7.5 & -5.8 & -4.9 \\ 
Qwen (ASR+audio, ctx=5) & 14.5 & 13.9 & 13.2 & 16.4 & 13.8 & 41.9 & -5.5 & -7.5 & -6.5 \\
    \bottomrule
    \end{tabular}
    \caption{Scores of all inspected XCOMET and Qwen variants. Averaged across all language directions, sorted by \kendall{} in descending order. Description of columns as in \Cref{tab:qwen-selection}.}
    \label{tab:qwen-avg-kendall}
\end{table*}